%% file: iclr2025_delta.tex
\documentclass{article} 
\usepackage{iclr2025_delta,times}

\input{math_commands.tex}

\usepackage{hyperref}
\usepackage{url}

\usepackage{algorithm}
\usepackage{algpseudocode}
\usepackage{mathtools}
\algnewcommand{\Initialize}[1]{
  \State \textbf{Initialize:}
  \State \hspace*{\algorithmicindent}\parbox[t]{0.8\linewidth}{\raggedright #1}
}

\usepackage{graphicx}
\usepackage{multirow}
\usepackage{makecell}
\usepackage{booktabs}

\definecolor{OI1}{RGB}{230,159,0}
\definecolor{OI2}{RGB}{0,158,115}
\definecolor{OI3}{RGB}{0,114,178}
\definecolor{OI4}{RGB}{213,94,0}
\definecolor{OI5}{RGB}{204,121,167}

\renewcommand{\eqref}[1]{(\ref{#1})}
\title{Image Interpolation with Score-based Riemannian Metrics of Diffusion Models}

\author{Shinnosuke Saito \& Takashi Matsubara \\
Hokkaido University \\
\texttt{\{saitou.shinnosuke.y0@elms, matsubara@ist\}.hokudai.ac.jp} \\
}

\iclrfinalcopy 

\begin{document}

\maketitle
\begin{abstract}
    Diffusion models excel in content generation by implicitly learning the data manifold, yet they lack a practical method to leverage this manifold---unlike other deep generative models equipped with latent spaces.
    This paper introduces a novel framework that treats the data space of pre-trained diffusion models as a Riemannian manifold, with a metric derived from the score function.
    Experiments with MNIST and Stable Diffusion show that this geometry-aware approach yields image interpolations that are more realistic, less noisy, and more faithful to prompts than existing methods, demonstrating its potential for improved content generation and editing.
\end{abstract}

\section{Introduction}
Deep generative models (DGMs) have achieved remarkable success in content generation across various domains \citep{ldm, sora, humanmotion, 3dobject}.
They also offer applications such as image attribute editing \citep{diffclip}, object replacement \citep{nullinv}, and smooth image interpolation \citep{noisediffusion}.
This success can be explained through the lens of the manifold hypothesis, which states that high-dimensional data lie on lower-dimensional manifolds.
DGMs equipped with latent spaces, including variational autoencoders (VAEs) \citep{vae} and generative adversarial networks (GANs) \citep{gan}, learn to model such manifolds by embedding latent spaces into data spaces \citep{bengiomfd, dgmdist, ganmfd, lowmfd, dgmhypothesis}.
Given this ability, recent work has explored insights from differential geometry, e.g., introducing Riemannian metrics to the latent spaces of pre-trained DGMs and generating semantically consistent content through traversals based on geodesics \citep{shaometric, chenmetric, latentodd, enrichlatent, bqriemannian, priorbased, clouddata}.

Diffusion models \citep{sohl, ddpm, ddim, scorebased}, a class of DGMs known for state-of-the-art generation quality, are also considered to learn data manifolds.
Some studies estimated their intrinsic dimensionality \citep{encodedim, kamkari, gaugedim, ventura}, and others improved the image quality by projecting samples onto the data manifolds during the generation \citep{invmfd, mpgd, stateguided}.
Nonetheless, they have not fully exploited the underlying Riemannian structures for tasks beyond naive sampling \citep{geodiff}.

In this paper, we introduce a Riemannian metric on the data space by leveraging the score function of diffusion models.
We examined a small model trained on MNIST \citep{mnist} and Stable Diffusion \citep{ldm} to demonstrate that our method yields more natural and faithful transitions, as assessed with CLIP-IQA \citep{Wang2023c}, compared with existing methods: linear (Lerp) \citep{ddpm} and spherical linear (Slerp) interpolation \citep{ddim}, as well as NAO \citep{nao} and Noise Diffusion \citep{noisediffusion}.
We also find that NAO and NoiseDiffusion suffer from severe reconstruction errors, whereas the other methods do not, as evaluated by mean squared error (MSE), LPIPS \citep{lpips}, and DreamSim \citep{dreamsim}.

\section{Related Work}

\paragraph{Deep Generative Models through Riemannian Geometry}
Pre-trained VAEs and GANs are known to be capable of editing generated image attributes (e.g., facial expressions and hair color) by linearly manipulating latent variables \citep{ganspace, voynov, zhu, shen, lowrankgan}.
To mitigate the constraint of being linear~\citep{latentodd}, some studies treated the latent space as a learned manifold with a Riemannian metric, improving the edit quality \citep{enrichlatent, bqriemannian, priorbased}.
A typical way defines a metric by pulling back the Euclidean metric from the data space, but it still assumed a linear data space, not fully capturing the underlying geometric structure \citep{shaometric, chenmetric, latentodd}.
Another way learns a metric on the data space and then pullback it to the latent space \citep{enrichlatent}, but it requires task-specific metric learning and additional architectures.
In diffusion models, the bottleneck layer of the U-Net for noise-prediction are suggested to serve as a latent space \citep{diffsemantic}.
Some studies assume a Euclidean metric on the bottleneck layer and pullback it to the data space, defining a metric on the data space \citep{geodiff, semanticdiff}.
Like the traditional ways for VAEs and GANs, this approach is however limited by the assumption of a linear latent space.

\paragraph{Data Manifolds of Diffusion Models}
Diffusion models are considered to learn data distributions $p_t(x_t)$ and the underlying manifolds at each diffusion time step $t$ \citep{detectmfd, dgmhypothesis}.
Several studies have estimated the local dimensionality of a data manifold $\gM_t$ in diffusion models by examining the trained score function $s_\theta(x_t,t)\approx\nabla_x\log p_t(x_t,t)$, specifically its Jacobian $\nabla_x s_\theta(x_t,t)$ (essentially the Hessian of $\log p_t(x_t,t)$) \citep{encodedim, ventura} or its divergence \citep{kamkari}.
Other works suggest that lower-quality samples arise when the reverse process drifts away from the manifold.
To mitigate this, some studies project the generated image $x_t$ onto the manifold $\gM_t$ by assuming a linear manifold \citep{invmfd}, by using a separate autoencoder \citep{mpgd}, or by constructing a subspace via singular value decomposition \citep{stateguided}.

Beyond naive sampling, some studies have tackled image interpolation \citep{ddpmzeroshot}.
Several methods require retraining \citep{diffmorpher, impus}, depend on specific models \citep{diffae, dbae, dblp}, or need additional conditioning \citep{wang}, and hence do not fully leverage the intrinsic manifold structure of pre-trained diffusion models.
The simplest method, Lerp \citep{ddpm}, linearly interpolates two images $x_t,x'_t$ after $t$ diffusion steps and then applies the reverse process.
This effectively treats the data space at time $t$ (often called a noise space) as a linear latent space, similar to VAEs or GANs.
Because the norm of a latent variable correlates with semantic richness \citep{nao,Alper2024}, Lerp leads to blur or loss of detail by decreasing norms.
Slerp \citep{slerp} preserves norms by interpolating along a spherical path.
NAO leverages the fact that norms of samples drawn from a normal distribution follow a chi distribution and maximizes the probability of the path between two endpoints \citep{nao}.
While these methods often yield smoother transitions than Lerp, they still lose detail and further generate artifacts because the latent variables for natural images often deviate from the expected normal distribution.
NoiseDiffusion addressed this by adding extra noise and clipping extreme noise~\citep{noisediffusion}, which indeed generates high-quality images but not necessarily a proper interpolation, as it can inject or remove information.
Ultimately, no existing interpolation method fully exploits the manifold structure in the data space.

\section{Method: Geodesic Interpolation}\label{sec:method}
\paragraph{Riemannian Metric based on Score Function}
For completeness, we provide background details on diffusion models and Riemannian geometry in Appendix~\ref{appendix:background}.
Here, we focus on our main proposal.
We propose the metric tensor $g$ in the data space $\gM_t$ at time $t$ of a diffusion model, which is represented by a matrix
\begin{equation}\label{eq:metric}
    \textstyle G_{x_t} = J_{x_t}^\top J_{x_t}\quad \text{for}\quad J_{x_t}=\nabla_{x_t} s_\theta(x_t, t).
\end{equation}
Since the score function $s_\theta(x_t, t)$ is an approximation of $\nabla_{x_t} \log p_t(x_t)$, its Jacobian $J_{x_t}$ corresponds to the Hessian $H_{x_t} = \nabla_{x_t} \nabla_{x_t} \log p_t(x_t)$.
As long as $J_{x_t}$ is non-degenerate, $G_{x_t}$ is positive definite and thus valid as a Riemannian metric.
With this metric, the length of a vector $v$ at $x_t$ is $|v|_g = \sqrt{\langle v,v\rangle_g}=\lVert J_{x_t}v\rVert_2$.
The length of a curve is obtained by integrating local lengths along the curve.
The length-minimizing curve between two points is called a \emph{geodesic}.
Thus, we define an interpolation between two samples $x_t^{\scriptscriptstyle (0)}$ and $x_t^{\scriptscriptstyle (N)}$ as the geodesic under the metric $g$ given by Eq.~(\ref{eq:metric}).

\paragraph{Interpretation of Proposed Metric}
The proposed metric is apparently similar to the pullback metric in prior works \citep{shaometric, chenmetric, latentodd,enrichlatent,geodiff, semanticdiff}.
However, these works use the Jacobian of a map from the data space to a latent space (or vice versa), whereas our metric employs the Jacobian of the score function and never pullbacks any predefined metric.
Note also that it is not the Hessian metric for a Hessian manifold.

In reality, observational noise prevents the data distribution from forming a perfectly low-dimensional manifold; the distribution is effectively collapsed (or compressed) along certain directions.
Since $s_\theta(x_t, t)$ is the gradient of the log-likelihood $\log p_t(x_t)$, it points in such directions to the manifold, guiding samples along directions of higher density.
Consequently, the directions corresponding to large eigenvalues of $J_{x_t}$ indicate collapsed dimensions, whereas those corresponding to small eigenvalues are tangential to the manifold \citep{encodedim, ventura}.
Therefore, following directions for which $\|J_{x_t}v\|$ is small can thus be seen as moving \emph{within} or \emph{parallel to} the manifold, providing a smooth transition of images.

Another interpretation follows from the Taylor expansion of $s_\theta$ around $x$, which yields $\lVert s_\theta(x + v, t) -s_\theta(x, t)\rVert_2= \lVert J_x v \rVert_2 + O(\lVert v\rVert_2^2)$.
This implies that the proposed geodesic corresponds to a curve along which $s_\theta(x, t)$ changes as little as possible.
Earlier studies have shown that the gradient of a log-likelihood (with respect to model parameters) can serve as a robust, semantically meaningful representation of input \citep{Charpiat2019,Hanawa2021,Yeh2018}.
In this light, our metric can be viewed as a measure of the \emph{semantic closeness} between infinitesimally different samples, providing transitions that preserve the underlying meaning within the data manifold.

\paragraph{Implementation}
Let $s \in [0,1]$ be the independent variable parameterizing a curve $\gamma: s(\in[0,1]) \mapsto \gamma(s)$.
The curve $\gamma$ is discretized as a sequence of data points $x_t^{\scriptscriptstyle (0)},\dots,x_t^{\scriptscriptstyle (N)}$, where $\gamma(s_i)=x^{\scriptscriptstyle (i)}$, $s_0=0$, $s_N=1$, and $s_{i+1}-s_i=\Delta s$.
The length of the curve, $L[\gamma]$, is numerically approximated using the trapezoidal rule:
\begin{equation}
    \textstyle
    L[\gamma] = \int_{0}^{1}l(s)\mathrm{d} s \approx \sum_{i=0}^{N-1}\frac{1}{2}(l(s_{i+1})+l(s_{i}))\Delta s,
\end{equation}
where the local path length $l(s_i)$ is given by $l(s_i)=\sqrt{\gamma'(s_i)^\top G_{\gamma(s_i)}\gamma'(s_i)}\approx\sqrt{(v_t^{\scriptscriptstyle (i)})^\top G_{x_t^{\scriptscriptstyle (i)}}v_t^{\scriptscriptstyle (i)}}$, and $v_t^{\scriptscriptstyle (i)}$ denotes the velocity at point $x_t^{\scriptscriptstyle (i)}$.
This is easily computed by the Jacobian-vector product.
The velocities $v_t^{\scriptscriptstyle (i)}$ are approximated using second-order finite differences.

Given two samples, $x_t^{\scriptscriptstyle (0)}$ and $x_t^{\scriptscriptstyle (N)}$, the geodesic path is obtained by minimizing the discrete approximation of $L[\gamma]$ with respect to the intermediate points $x_t^{\scriptscriptstyle (1)}, \dots, x_t^{\scriptscriptstyle (N-1)}$, i.e.,
\begin{equation}
    \textstyle \min_{x_t^{\scriptscriptstyle (1)},..,x_t^{\scriptscriptstyle (N-1)}} L[\gamma]
    \quad \textrm{s.t.}~\gamma(s_{0})=x_t^{\scriptscriptstyle (0)},\gamma(s_{N})=x_t^{\scriptscriptstyle (N)}.
\end{equation}
To prevent the intermediate points from collapsing to a single point, we add the variance of the Euclidean distance, $\mathrm{Var}\!\left[\lVert x_t^{\scriptscriptstyle (i+1)} - x_t^{\scriptscriptstyle (i)}\rVert_2\right]$, as a regularization term to the loss function, multiplied by hyperparameter $\lambda$.
This term guides the velocity to be constant.
The initial values of $x_t^{\scriptscriptstyle (1)}, \dots, x_t^{\scriptscriptstyle (N-1)}$ can be initialized using any reasonable method; in this work, we employed Slerp \citep{ddim}.

We observed that directly computing the geodesic in the raw data space at $t = 0$ leads to poor results due to the highly rugged landscape of the score function $s_\theta$ at $t = 0$, likely because it memorizes individual training data points.
To address this issue, we employed DDIM inversion to map samples into the data space at a specific time step $t = \tau>0$, compute the geodesic there, and then apply the reverse DDIM process to obtain the final image sequence in the original data space at $t=0$.

\section{Experiments and Results}

We evaluated our method with Stable Diffusion \citep{ldm}, as well as Lerp \citep{ddpm}, Slerp \citep{slerp}, NAO \citep{nao}, and NoiseDiffusion \citep{noisediffusion}.
We set max forward steps to $T=50$ and the number of forward steps before interpolation to $\tau=50$ for NAO and $30$ for others.
We set the number of interpolation steps to $N=10$.
We optimized the path using Adam \citep{adam} for 5,000 iterations.
The learning rate was initialized at $10^{-2}$ and decayed to zero using cosine annealing~\citep{Loshchilov2017}.

\begin{figure}[t]
    \centering
    \scriptsize
    \setlength{\tabcolsep}{0.5mm}
    \begin{tabular}{rrr}
                                                     & \hspace*{-1cm}\includegraphics[width=6.3cm]{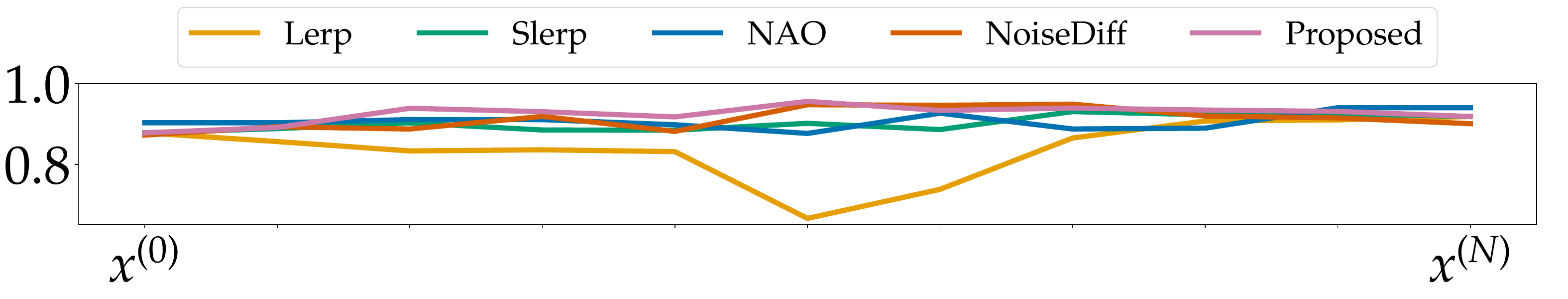}                                     & \hspace*{-1cm}\includegraphics[width=6.3cm]{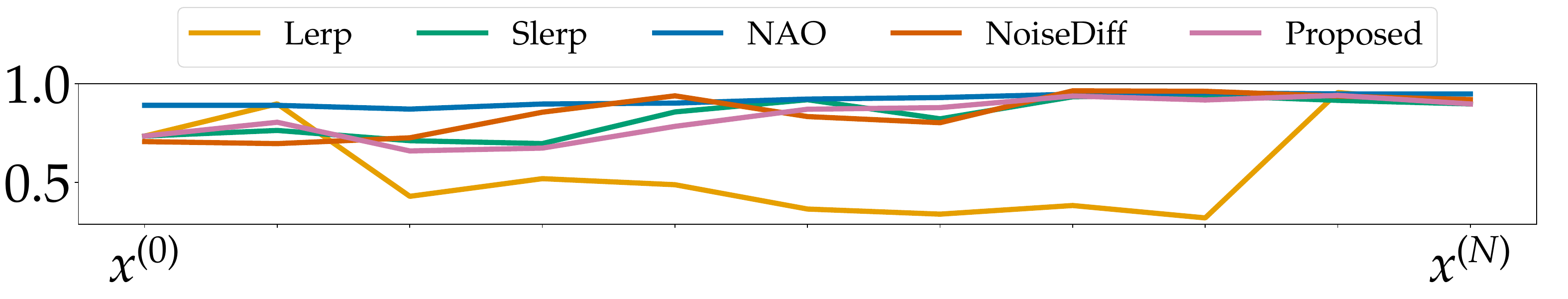}                                       \\
        \raisebox{2.5mm}{\textcolor{OI1}{Lerp}}      & {\includegraphics[width=6.3cm]{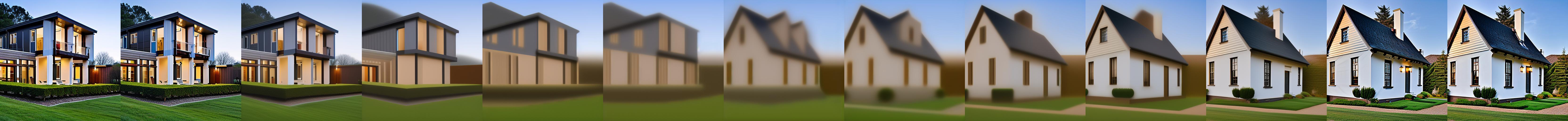}}                                                & {\includegraphics[width=6.3cm]{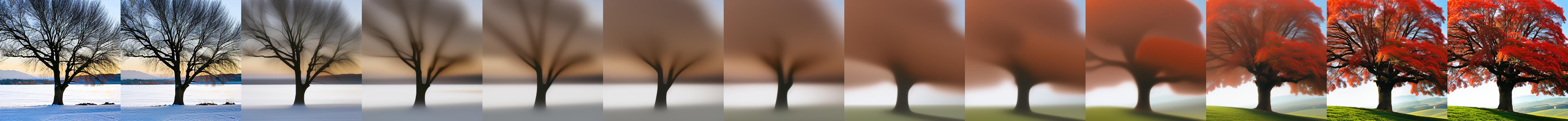}}                                                  \\[-0.5mm]
        \raisebox{2.5mm}{\textcolor{OI2}{Slerp}}     & {\includegraphics[width=6.3cm]{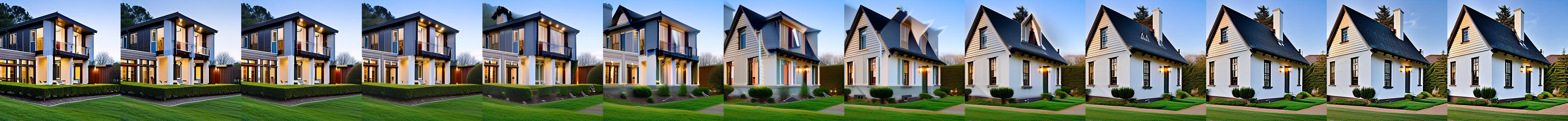}}                                               & {\includegraphics[width=6.3cm]{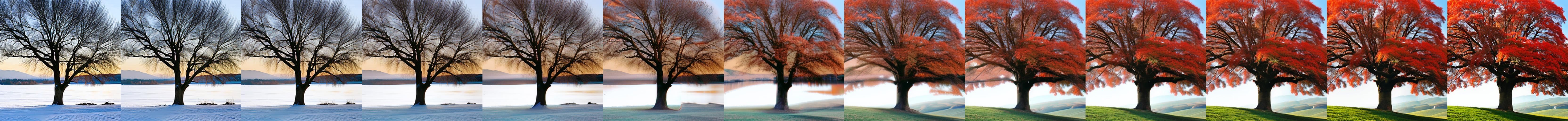}}                                                 \\[-0.5mm]
        \raisebox{2.5mm}{\textcolor{OI4}{NAO}}       & {\includegraphics[width=6.3cm]{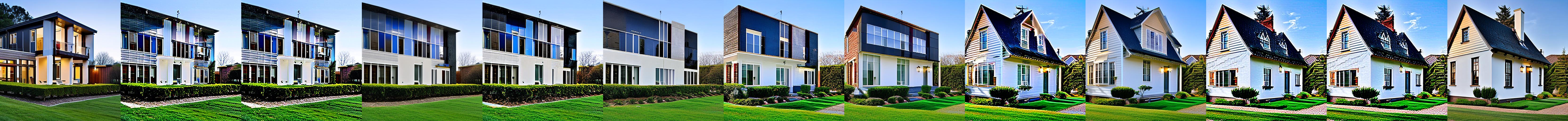}}                                                 & {\includegraphics[width=6.3cm]{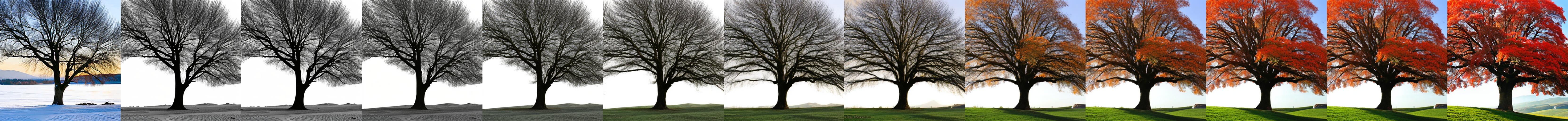}}                                                   \\[-0.5mm]
        \raisebox{2.5mm}{\textcolor{OI3}{NoiseDiff}} & {\includegraphics[width=6.3cm]{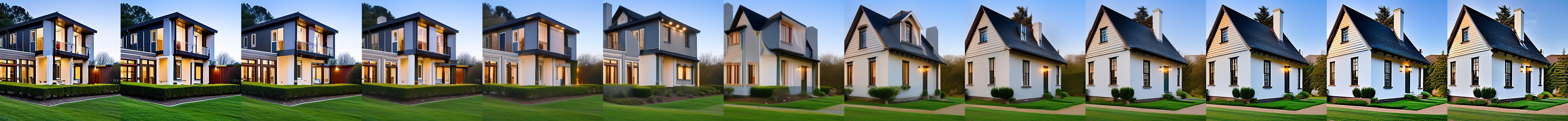}}                                                  & {\includegraphics[width=6.3cm]{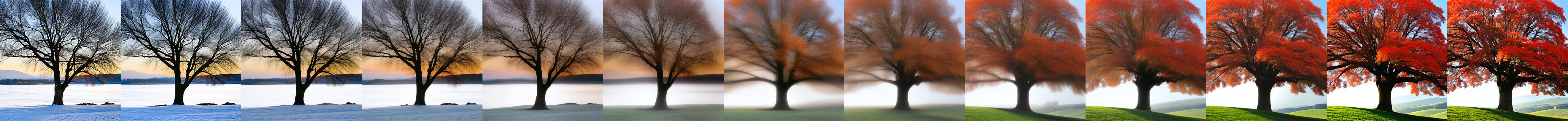}}                                                    \\[-0.5mm]
        \raisebox{2.5mm}{\textcolor{OI5}{proposed}}  & {\includegraphics[width=6.3cm]{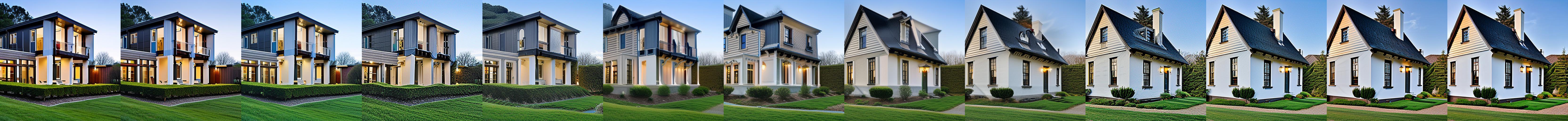}}                                                 & {\includegraphics[width=6.3cm]{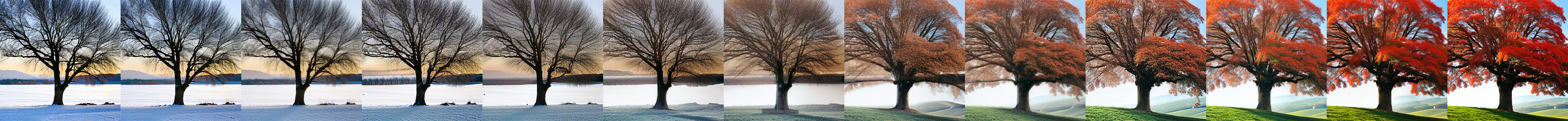}}                                                   \\[1.0mm]
                                                     & org. \ $x^{\scriptscriptstyle (0)}$ \hfill ``a photo of a house'' \hfill   $x^{\scriptscriptstyle (N)}$ org. & org. \ $x^{\scriptscriptstyle (0)}$ \hfill ``a photo of a tree'' \hfill   $x^{\scriptscriptstyle (N)}$ org.  \\[0.5mm]
                                                     & \hspace*{-1cm}\includegraphics[width=6.3cm]{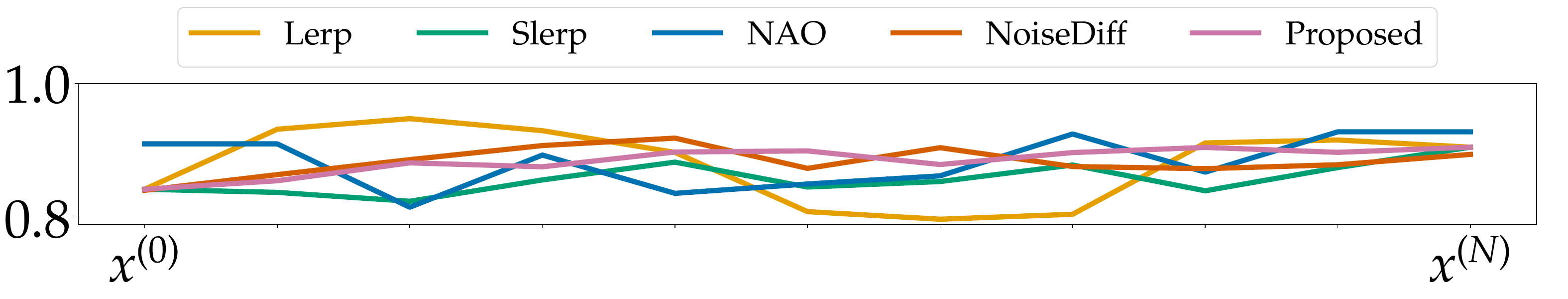}                                         & \hspace*{-1cm}\includegraphics[width=6.3cm]{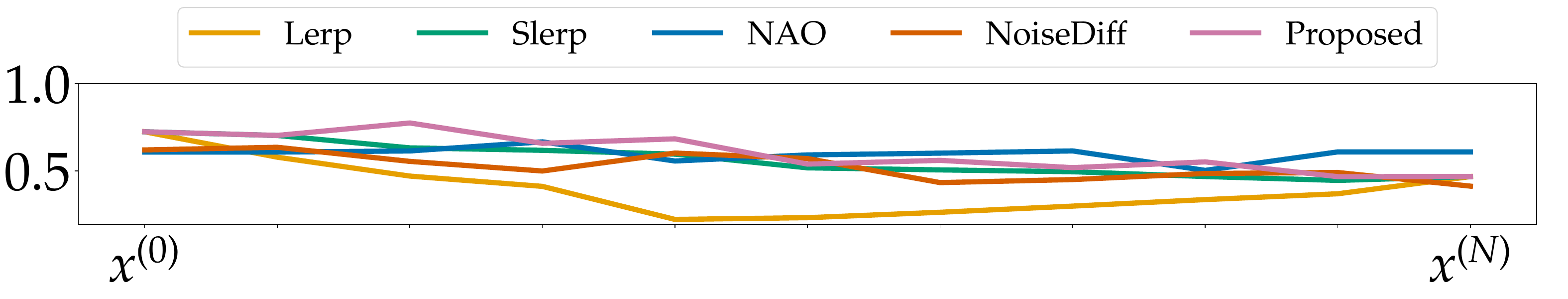}                                       \\
        \raisebox{2.5mm}{\textcolor{OI1}{Lerp}}      & {\includegraphics[width=6.3cm]{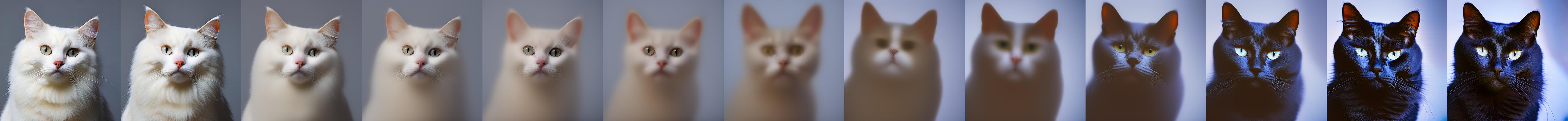}}                                                    & {\includegraphics[width=6.3cm]{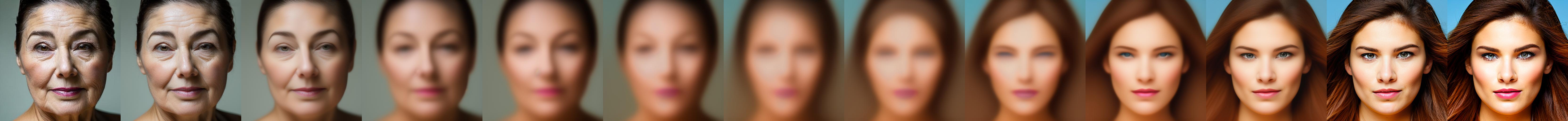}}                                                  \\[-0.5mm]
        \raisebox{2.5mm}{\textcolor{OI2}{Slerp}}     & {\includegraphics[width=6.3cm]{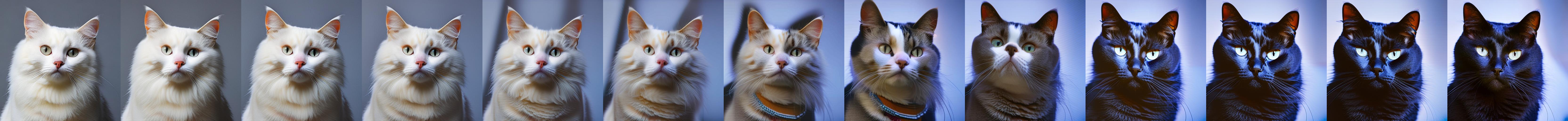}}                                                   & {\includegraphics[width=6.3cm]{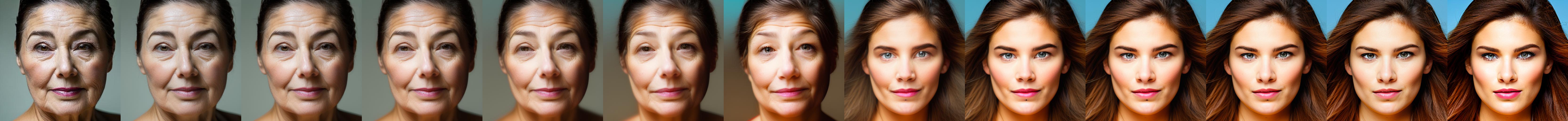}}                                                 \\[-0.5mm]
        \raisebox{2.5mm}{\textcolor{OI4}{NAO}}       & {\includegraphics[width=6.3cm]{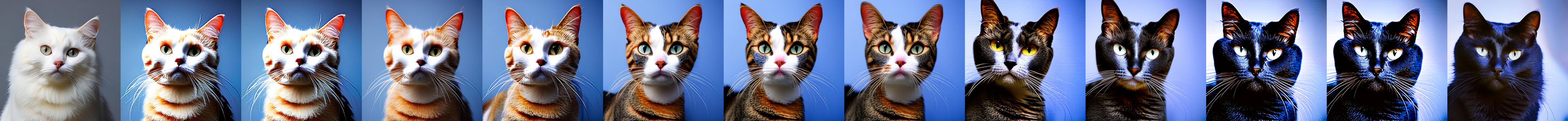}}                                                     & {\includegraphics[width=6.3cm]{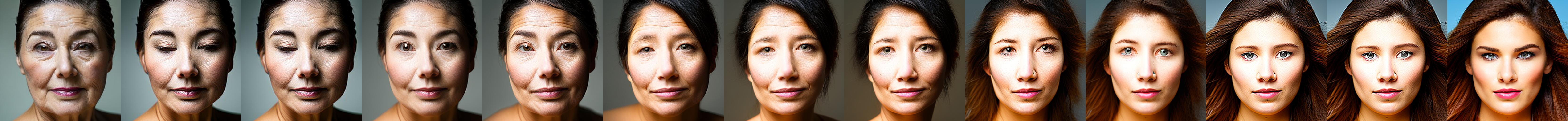}}                                                   \\[-0.5mm]
        \raisebox{2.5mm}{\textcolor{OI3}{NoiseDiff}} & {\includegraphics[width=6.3cm]{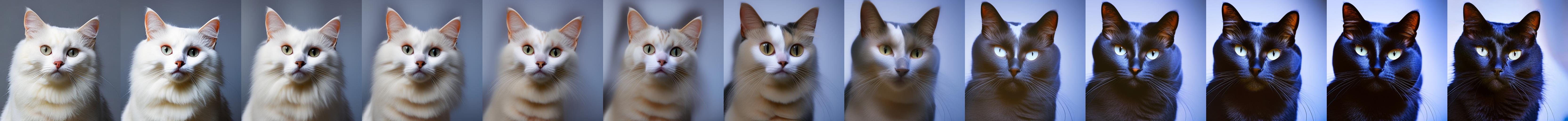}}                                                      & {\includegraphics[width=6.3cm]{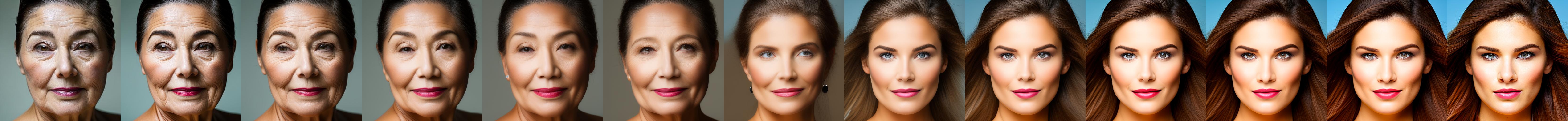}}                                                    \\[-0.5mm]
        \raisebox{2.5mm}{\textcolor{OI5}{proposed}}  & {\includegraphics[width=6.3cm]{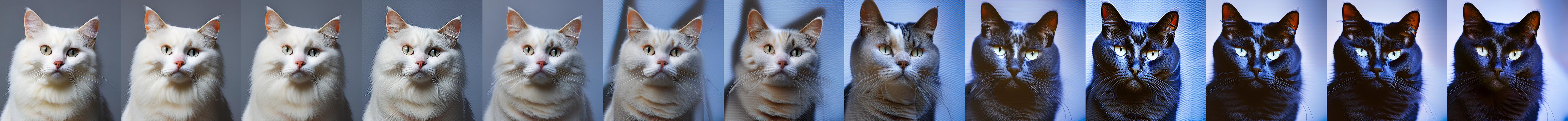}}                                                     & {\includegraphics[width=6.3cm]{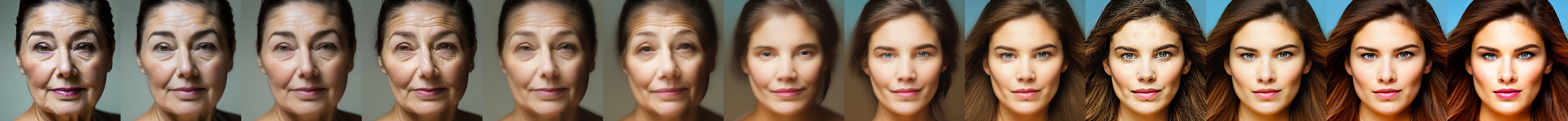}}                                                   \\
                                                     & org. \ $x^{\scriptscriptstyle (0)}$ \hfill ``a photo of a cat'' \hfill   $x^{\scriptscriptstyle (N)}$ org.   & org. \ $x^{\scriptscriptstyle (0)}$ \hfill ``a photo of a woman'' \hfill   $x^{\scriptscriptstyle (N)}$ org. \\[-2mm]
    \end{tabular}
    \caption{Images generated by Stable Diffusion with prompts shown below, with interpolations using different methods (CFG: 7.5).
        Images at both ends are original, adjacent to them are the reconstructions, and in between are the interpolation results.
        The sample-wise fidelity is visualized above each image as a graph.
    }
    \label{fig:sd}
\end{figure}

Figure~\ref{fig:sd} and Table~\ref{tb:rec_error} summarize the results.
As previous works have shown, Lerp produces blurry images with a notable loss of detail.
Slerp yields smoother transitions, but sometimes objects appear doubled, like houses or trees.
NAO and NoiseDiffusion suffer from severe reconstruction errors because of a long diffusion process and added or clipped noise, as evaluated by MSE, LPIPS \citep{lpips}, and DreamSim \citep{dreamsim}.
In contrast, the proposed method demonstrates the most realistic and faithful transitions, gradually adjusting objects, color, and lighting while avoiding noise, as assessed with CLIP-IQA.
See Appendix~\ref{appendix:results} for additional explanations and results.

\vspace{\baselineskip}
\begin{table}[t]
    \centering
    \scriptsize
    \caption{Mean Reconstruction Errors and CLIP-IQA Assessment for Four Examples in Figure~\ref{fig:sd}}
    \vspace{\baselineskip}
    \label{tb:rec_error}
    \begin{tabular}{lcccccc}
        \toprule
                        & \multicolumn{3}{c}{Reconstruction Errors} & \multicolumn{3}{c}{CLIP-IQA}                                                                                                          \\
        \cmidrule(lr){2-4} \cmidrule(lr){5-7}
        \textbf{Method} & \textbf{MSE $[\times 10^{-3}]$}           & \textbf{LPIPS $[\times 10^{-1}]$} & \textbf{DreamSim $[\times 10^{-2}]$} & \textbf{Reality}  & \textbf{Noisiness} & \textbf{Fidelity} \\
        \midrule
        Lerp            & \textbf{\ \ 7.89}                         & \textbf{1.71}                     & \textbf{\ \ 5.34}                    & 0.389             & 0.406              & 0.686             \\
        Slerp           & \textbf{\ \ 7.89}                         & \textbf{1.71}                     & \textbf{\ \ 5.34}                    & \underline{0.704} & 0.765              & 0.784             \\
        NAO             & 64.04                                     & 6.03                              & 32.82                                & 0.617             & \underline{0.766}  & \textbf{0.815}    \\
        NoiseDiff       & 14.13                                     & 2.35                              & \ \ 6.93                             & 0.600             & 0.646              & 0.783             \\
        Proposed        & \textbf{\ \ 7.89}                         & \textbf{1.71}                     & \textbf{\ \ 5.34}                    & \textbf{0.716}    & \textbf{0.818}     & \underline{0.810} \\
        \bottomrule
    \end{tabular}
\end{table}

\section{Conclusion}
This paper introduces a Riemannian metric derived from the score function of pre-trained diffusion models.
The metric yields geodesic paths that naturally follow the learned data manifold, providing a geometry-aware framework for image interpolation.
Experiments on MNIST and Stable Diffusion show smooth, natural, and faithful transitions than existing methods.
The proposed geometric framework has broader potential applications, such as video editing by treating a video as a curve on the manifold and using parallel transport to modify all frames simultaneously.

\clearpage

\bibliography{iclr2025_delta}
\bibliographystyle{iclr2025_delta}

\appendix
\clearpage

\section{Background Theory}\label{appendix:background}
This section provides the foundational concepts necessary for our proposed method.
We begin with an overview of diffusion models, describing their forward and reverse processes and the link between noise prediction and the score function.
We then introduce key elements of Riemannian geometry, focusing on how Riemannian metrics induce distances and paths on manifolds.

\subsection{Diffusion Models}
Diffusion models are a class of DGMs inspired by non-equilibrium thermodynamics \citep{sohl, ddpm}.
The model consists of two processes: a forward process that adds noise to the data and a reverse process that removes the noise.
Their core idea is to model the underlying data distribution by denoising noisy samples.

\paragraph{Forward Process}
Let $x_0 \in \mathbb{R}^D$ be a data sample.
The forward process is defined as a Markov chain in which Gaussian noise is added at each time step $t=1,\dots,T$:
\begin{equation}
    \label{forward}
    q({x_t}|{x}_{t-1})=\mathcal{N}\left({x}_t;\sqrt{1-\beta_t}{x}_{t-1},\beta_t {{I}}\right)=\mathcal{N}\left(\sqrt{\frac{\alpha_t}{\alpha_{t-1}}}{x}_{t-1}, \left(1-\frac{\alpha_t}{\alpha_{t-1}}\right){{I}}\right),
\end{equation}
where $\{\beta_t\}_{t=1}^T$ is a variance schedule, $I$ is the identity matrix in $\mathbb{R}^D$, and $\alpha_t=\prod_{s=1}^t(1-\beta_s)$.
As $t$ increases, $x_t$ becomes progressively more corrupted by noise until $x_T$ is nearly an isotropic Gaussian distribution.

\paragraph{Reverse Process}
To invert this forward process, a reverse Markov chain $p_\theta({x}_{t-1}|{x}_t)$ from ${x}_T\sim\mathcal{N}(0,{I})$ is constructed as
\begin{equation}
    \label{backward}
    {x}_{t-1}=\frac{1}{\sqrt{1-\beta_t}}\left({x_t}-\frac{\beta_t}{\sqrt{1-\alpha_t}}{\epsilon}_\theta({x}_t, t)\right)+\sigma_t{z}_t,
\end{equation}
with trainable noise predictor ${\epsilon}_\theta$, where ${z}_t \sim \mathcal{N}(0,{I})$ and $\sigma_t^2 = \beta_t$ is a variance.
The noise predictor ${\epsilon}_\theta({x}_t,t)$ is trained by minimizing the objective:
\begin{equation}
    \label{ddpm_obj}
    \mathcal{L}(\theta)=\mathbb{E}_{{x},{\epsilon}_t, t}\big[\|{\epsilon}_t-{\epsilon}_\theta({x}_t, t)\|_2^2\big],
\end{equation}
where ${\epsilon}_t\sim\mathcal{N}(0,{I})$ is the noise added during the forward process at time step $t$.

\paragraph{Denoising Diffusion Implicit Models}
Denoising diffusion implicit models (DDIMs) \citep{ddim} modified Eq.~(\ref{forward}) as a non-Markovian process $q({x}_{t-1}|{x}_t,{x}_0)=\mathcal{N}(\sqrt{\vphantom{/}\alpha_{t-1}}{x}_{0}+\sqrt{1-\alpha_{t-1}-\sigma_t^2},\sigma_t^2 {{I}})$.
Then the reverse process becomes
\begin{equation}
    \label{backward_ddim}
    {x}_{t-1}=\sqrt{\alpha_{t-1}}\left(\frac{{x_t}-\sqrt{1-\alpha_{t}}{\epsilon}_\theta({x}_t,t)}{\sqrt{\alpha_t}}\right)+\sqrt{1-\alpha_{t-1}-\sigma_t^2}\cdot{\epsilon}_\theta({x}_t,t)+\sigma_t{z}_t,
\end{equation}
where $\sigma_t=\eta\sqrt{(1-\alpha_{t-1})/(1-\alpha_t)}\sqrt{1-\alpha_t/\alpha_{t-1}}$.
Here, $\eta\in[0,1]$ determines the stochasticity: $\eta=1$ recovers the DDPM, while $\eta=0$ yields a deterministic update.

\paragraph{Formulation as Stochastic Differential Equations}
Diffusion models can also be formulated using stochastic differential equations (SDEs) \citep{scorebased}.
In that viewpoint, the forward process is governed by a continuous-time SDE, and its time-reversal is defined through the corresponding reverse-time SDE, which depends on the score function $\nabla_{x_t}\log p_t(x_t)$, where $p_t(\cdot)$ denotes the distribution of $x_t$ at time $t$.
Notably, the noise-prediction network $\epsilon_\theta$ is closely tied to the score function \citep{unified} as:
\begin{equation}
    \label{score_noise_relation}
    \nabla_{{x}_t} \log p_t({x}_t) \approx -\frac{1}{\sqrt{1-\alpha_t}}{\epsilon}_\theta({x}_t, t).
\end{equation}
Hence, learning $\epsilon_\theta$ for noise prediction is equivalent to learning the score function.

\subsection{Riemannian Geometry}

We adopt the presentation in \citet{rmlee2019}.
Let $\gM$ be a smooth manifold.
A \emph{Riemannian metric} $g$ on $\gM$ is a smooth $(0,2)$-tensor field such that at every point $p \in \gM$, the tensor $g_p$ defines an inner product on the tangent space $T_p\gM$.
Concretely, $g_p$ is symmetric and positive-definite:
\begin{equation*}
    g_p(v,v) \ge 0 \quad\text{for all }v \in T_p\gM\quad \text{and}\quad
    g_p(v,v) = 0 \Leftrightarrow v=0.
\end{equation*}
By identifying $g_p$ with an inner product, we write
\begin{equation*}
    \langle v, w\rangle_g := g_p(v,w) \quad \text{for }v,w\in T_p\gM.
\end{equation*}
A \emph{Riemannian manifold} is then the pair $(\gM, g)$.

Let $(x^1,\dots,x^n)$ be local coordinates near $p\in\gM$.
Then, the basis for $T_p\gM$ is $\bigl(\frac{\partial}{\partial x^1}\big|_p,\dots,\frac{\partial}{\partial x^n}\big|_p\bigr)$.
Tangent vectors $v,w\in T_p{\gM}$ are expressed as $v=\sum_{i=1}^nv^i\left.\frac{\partial}{\partial x^i}\right|_{p}$ and $w=\sum_{i=1}^nw^i\left.\frac{\partial}{\partial x^i}\right|_{p}$, respectively.
The matrix notation $G_p$ of $g$ at $p$ consists of $(i,j)$-elements
\begin{equation}
    \label{metric tensor}
    g_{ij}(p)=g_p\left(\left.\frac{\partial}{\partial x^i}\right|_{p},\left.\frac{\partial}{\partial x^j}\right|_{p}\right)=\left\langle \left.\frac{\partial}{\partial x^i}\right|_{p},\left.\frac{\partial}{\partial x^j}\right|_{p} \right\rangle_g
\end{equation}
for $i,j=1,2,...,n$.
The Euclidean metric is written as an identity matrix $I$.
The inner product of $v$ and $w$ with respect to the Riemannian metric $g_p$ can be expressed as:
\begin{equation}
    \label{inner product with metric tensor}
    g_p(v,w)=\sum_{i,j=1}^ng_{ij}(p)v^iw^j=v^TG_pw.
\end{equation}

\paragraph{Lengths of Tangent Vectors and Curves.}
Given $\langle v,v\rangle_g$, the length of a tangent vector $v\in T_p\gM$ is
given by $|v|_g := \sqrt{\langle v,v\rangle_g}$.
For a smooth curve $\gamma:[0,1]\to \gM$, its length is defined by
\begin{equation}
    \label{length}
    L(\gamma):=\int_0^1|\gamma'(t)|_gdt=\int_0^1\sqrt{\langle \gamma'(t),\gamma'(t)\rangle_g}dt = \int_0^1 \sqrt{\gamma'(t)^\top G_{\gamma(t)} \gamma'(t)} dt.
\end{equation}
For convenience, we denote this integrand of Eq.~(\ref{length}) as:
\begin{equation}
    \label{integrad}
    l(t):=\sqrt{\gamma'(t)^\top G_{\gamma(t)} \gamma'(t)}.
\end{equation}

\section{Comparison Methods}

\subsection{DDIM Inversion}
Naive encoding of an original image is simply adding Gaussian noise as in the forward process $q({x_t}|{x}_{t-1})$, which is stochastic and often yields poor reconstructions.
To accurately invert the reverse process and recover the specific noise map associated with a given image, \emph{DDIM Inversion} \citep{nullinv} is widely used.
The key insight is that, in the limit of infinitesimally small time steps, the ODE formulation of DDIM is invertible.

Concretely, setting $\sigma_t=0$ in Eq.~(\ref{backward_ddim}) gives
\begin{equation}
    \label{ddim_update_abb}
    x_{t-1} = a_t x_t + b_t \epsilon_\theta(x_t, t),
\end{equation}
where $a_t=\sqrt{\alpha_{t-1}/\alpha_t}$ and $b_t=-\sqrt{\alpha_{t-1}(1-\alpha_t)/\alpha_t}+\sqrt{\vphantom{/}1-\alpha_{t-1}}$.
With a sufficiently small time step size,
\begin{equation}
    \label{inversion}
    x_t = \frac{x_{t-1} - b_t \epsilon_\theta(x_t, t)}{a_t} \approx \frac{x_{t-1} - b_t \epsilon_\theta(x_{t-1}, t)}{a_t},
\end{equation}
as $\epsilon_\theta(x_{t},t)\approx\epsilon_\theta(x_{t-1},t)$.
Iteratively applying the update rule in Eq.~(\ref{inversion}) to a sample $x_0$ from $t=1$ to $\tau$ recovers the noisy image $x_\tau$ that would generate the original $x_0$.
This inversion procedure substantially improves the fidelity of reconstructions and subsequent interpolations.

\subsection{Linear Interpolation}
Once the noisy images are recovered via DDIM Inversion, one can perform straightforward linear interpolation (Lerp) \citep{ddim}, by treating the noise space (the data space with $t>0$) as a linear latent space.
In particular, let $x_\tau^{\scriptscriptstyle (0)}$ and $x_\tau^{\scriptscriptstyle (1)}$ denote the noisy versions of $x_0^{\scriptscriptstyle (0)}$ and $x_0^{\scriptscriptstyle (1)}$ in the noise space at $t=\tau$, respectively.
A linear interpolation in that space is given by
\begin{equation}
    {x}_\tau^{\scriptscriptstyle (s)}=(1-s)x_\tau^{\scriptscriptstyle (0)}+s x_\tau^{\scriptscriptstyle (1)},
\end{equation}
where $s\in[0,1]$ is the interpolation parameter.
Then, one then applies the reverse process from $t=\tau$ back to $t=0$ to obtain the interpolated images $x_0^{\scriptscriptstyle (s)}$ in the data space.

\subsection{Spherical Linear Interpolation}
An alternative is spherical linear interpolation (Slerp) \citep{slerp}, which finds the shortest path on the unit sphere in the noise space:
\begin{equation}
    x_\tau^{\scriptscriptstyle (s)}=\frac{\sin{((1-s)\theta})}{\sin{(\theta)}}x_\tau^{\scriptscriptstyle (0)}+\frac{\sin{(s\theta})}{\sin{(\theta)}}x_\tau^{\scriptscriptstyle (1)}
\end{equation}
where $\theta=\arccos\left({\frac{(x_\tau^{\scriptscriptstyle (0)})^\top x_\tau^{\scriptscriptstyle (1)}}{\|x_\tau^{\scriptscriptstyle (0)}\| \|x_\tau^{\scriptscriptstyle (0)}\|}}\right)$.
Because this procedure preserves the norms of the noisy images $x_\tau^{\scriptscriptstyle (s)}$, it often yields natural interpolations than Lerp.
Note that, Slerp assumes that $x_\tau^{\scriptscriptstyle (0)}$ and $x_\tau^{\scriptscriptstyle (1)}$ are drawn from a normal distribution, which holds only for a sufficiently large $\tau$.
Nonetheless, Slerp typically performs better with moderate $\tau$.

\subsection{Proposed Method}
The velocities $v_t^{\scriptscriptstyle (i)}$ are approximated using second-order finite differences:
\begin{equation}
    v_t^{\scriptscriptstyle (i)} =
    \begin{cases}
        \frac{-3x_t^{\scriptscriptstyle (0)} + 4x_t^{\scriptscriptstyle (1)} - x_t^{\scriptscriptstyle (2)}}{2\Delta s}    & (i = 0)     \\
        \frac{x_t^{\scriptscriptstyle (i+1)} - x_t^{\scriptscriptstyle (i-1)}}{2\Delta s}                                  & (0 < i < N) \\
        \frac{3x_t^{\scriptscriptstyle (N)} - 4x_t^{\scriptscriptstyle (N-1)} + x_t^{\scriptscriptstyle (N-2)}}{2\Delta s} & (i = N)
    \end{cases}
\end{equation}

\section{Additional Experiments and Results}\label{appendix:results}
\subsection{CLIP-IQA}
CLIP-IQA \citep{Wang2023c} is a metric that evaluates the quality of images generated by generative models.
It is based on a pre-trained language-image model, CLIP \citep{Radford2021}, which predicts the similarity between images and text.
For evaluating the reality of images, CLIP-IQA uses a pair of prompts: ``Real photo'' and ``Abstract photo,'' and evaluate how similar the generated images are to the prompts.
If it is close to the ``Real photo'' prompt, the score approaches 1.0, and the image is considered realistic.
For noisiness, the prompts are ``Clean photo'' and ``Noisy photo''.

While this goes beyond CLIP-IQA’s original scope, we used the prompts “A photo of [object]” and “A photo of something that is not [object]” to evaluate the fidelity of interpolated images to the prompt.

\subsection{Experiments with MNIST}

We also evaluated our method with MNIST \citep{mnist}, as well as Lerp \citep{ddim}, Slerp \citep{slerp}.
We set the maximum number of forward steps to $T=1000$ and the number of forward steps before interpolation to $\tau=400$.
We set the number of interpolation steps to $N=10$.
We optimized the geodesic path using Adam \citep{adam} for 5,000 iterations.
The learning rate was initialized at $10^{-3}$.

\begin{figure}[t]
    \centering
    \scriptsize
    \setlength{\tabcolsep}{0.5mm}
    \begin{tabular}{rrr}
        \raisebox{4mm}{\textcolor{OI1}{Lerp}}     & {\includegraphics[width=6.5cm]{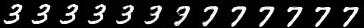}}                                                                                   \\[-0.5mm]
        \raisebox{4mm}{\textcolor{OI2}{Slerp}}    & {\includegraphics[width=6.5cm]{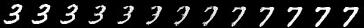}}                                                                                  \\[-0.5mm]
        \raisebox{4mm}{\textcolor{OI5}{Proposed}} & {\includegraphics[width=6.5cm]{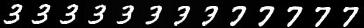}}                                                                                    \\[1mm]
        \raisebox{4mm}{\textcolor{OI1}{Lerp}}     & {\includegraphics[width=6.5cm]{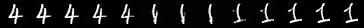}}                                                                                   \\[-0.5mm]
        \raisebox{4mm}{\textcolor{OI2}{Slerp}}    & {\includegraphics[width=6.5cm]{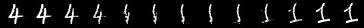}}                                                                                  \\[-0.5mm]
        \raisebox{4mm}{\textcolor{OI5}{Proposed}} & {\includegraphics[width=6.5cm]{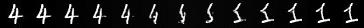}}                                                                                    \\[1mm]
        \raisebox{4mm}{\textcolor{OI1}{Lerp}}     & {\includegraphics[width=6.5cm]{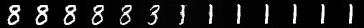}}                                                                                   \\[-0.5mm]
        \raisebox{4mm}{\textcolor{OI2}{Slerp}}    & {\includegraphics[width=6.5cm]{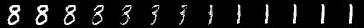}}                                                                                  \\[-0.5mm]
        \raisebox{4mm}{\textcolor{OI5}{Proposed}} & {\includegraphics[width=6.5cm]{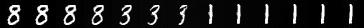}}                                                                                    \\[-0.5mm]
                                                  & \hspace{1.25mm} org. \hspace{1.25mm} $x^{\scriptscriptstyle (0)}$ \hfill $x^{\scriptscriptstyle (N)}$ \hspace{1.25mm} org. \hspace{1.25mm} \\[-2mm]
    \end{tabular}
    \caption{Interpolation results by a diffusion model trained on MNIST.}
    \label{fig:mnist}
\end{figure}

Unlike the results from Stable Diffusion (Figure \ref{fig:sd}), Lerp produces interpolated results with less noisy than Slerp.
This suggests that the MNIST data manifold is relatively locally linear.
However, Lerp exhibits discontinuous changes in digits; for example, a sudden appearance of a ``9'' during the transition from ``3'' to ``7''.
This is likely because Lerp ignores the underlying metric of the data manifold.
In contrast, the proposed method shows gradual transitions compared to both Lerp and Slerp, achieving geometrically consistent transitions.

\end{document}

%% file: math_commands.tex
\usepackage{amsmath,amsfonts,bm}

\def\eqref#1{equation~\ref{#1}}

\def\1{\bm{1}}

\DeclareMathAlphabet{\mathsfit}{\encodingdefault}{\sfdefault}{m}{sl}
\SetMathAlphabet{\mathsfit}{bold}{\encodingdefault}{\sfdefault}{bx}{n}

\def\gM{{\mathcal{M}}}

